\documentclass{article} 
\usepackage{latex_template,times}
\usepackage{hyperref}
\usepackage{amsmath,amsfonts,amssymb}
\usepackage{algorithm}
\usepackage{algpseudocode}
\usepackage{url}
\usepackage{listings}
\usepackage{changepage}
\usepackage{float}
\usepackage{caption}
\usepackage{graphicx}
\usepackage{booktabs}
\usepackage{amsmath}
\usepackage{array}
\usepackage{appendix}
\usepackage{listings}
\usepackage{multirow}
\usepackage{subfigure}
\usepackage{arydshln}

\algnewcommand\INPUT{\item[\textbf{Input:}]}%
\algnewcommand\OUTPUT{\item[\textbf{Output:}]}%
\algnewcommand\STOP{\item[\textbf{Stop:}]}%
\newcolumntype{P}[1]{>{\centering\arraybackslash}p{#1}}

\newtheorem{theorem}{Theorem}

\title{Group Reasoning Emission Estimation Networks}

\nipsfinalcopy 

\begin{document}

\author{
  Yanming Guo \\
  University of Sydney \\
  \texttt{yguo0337@uni.sydney.edu.au} \\
  \And
  Qian Xiao \\
  Maynooth University \\
  \texttt{xiaoqian.kasey@outlook.com} \\
  \And
  Kevin Credit \\
  Maynooth University \\
  \texttt{kevin.credit@mu.ie} \\
  \And
  Jin Ma \\
  University of Sydney \\
  \texttt{j.ma@sydney.edu.au} \\
}

\maketitle

\begin{abstract}
Accurate greenhouse gas (GHG) emission reporting is critical for governments, businesses, and investors. However, adoption remains limited—particularly among small and medium enterprises—due to high implementation costs, fragmented emission factor databases, and a lack of robust sector classification methods. To address these challenges, we introduce \textbf{Group Reasoning Emission Estimation Networks (GREEN)}, an AI-driven carbon accounting framework that standardizes enterprise-level emission estimation, constructs a large-scale benchmark dataset, and leverages a novel reasoning approach with large language models (LLMs). Specifically, we compile textual descriptions for 20,850 companies with validated North American Industry Classification System (NAICS) labels and align these with an economic model of carbon intensity factors. By reframing sector classification as an information retrieval task, we fine-tune Sentence-BERT models using a contrastive learning loss. To overcome the limitations of single-stage models in handling thousands of hierarchical categories, we propose a \emph{Group Reasoning} method that ensembles LLM classifiers based on the natural NAICS ontology, decomposing the task into multiple sub-classification steps. We theoretically prove that this approach reduces classification uncertainty and computational complexity. Experiments on 1,114 NAICS categories yield state-of-the-art performance (83.68\% Top-1, 91.47\% Top-10 accuracy), and case studies on 20 companies report a mean absolute percentage error (MAPE) of 45.88\%. The project is available at: \url{https://huggingface.co/datasets/Yvnminc/ExioNAICS}.
\end{abstract}

\section{Introduction}
\label{sec:intro}
Accurate greenhouse gas (GHG) emission reporting has become increasingly critical for governments, businesses, and investors striving to mitigate the impacts of climate change~\cite{emission1,emission2}. In recent years, various jurisdictions worldwide have instituted mandatory disclosure frameworks that oblige enterprises to publicly report their emissions, spurring both transparency and accountability in corporate climate action~\cite{esg, gov}. For instance, the European Union Emissions Trading System (EU ETS) imposes stringent reporting and trading requirements on power and industrial sectors, contributing to notable emission reductions since its inception~\cite{eureport}. Australia, through the National Greenhouse and Energy Reporting (NGER) Act 2007, similarly enforces comprehensive disclosure obligations designed to capture reliable emissions data from large facilities~\cite{aureport}. Beyond regulatory obligations, investors and other stakeholders increasingly demand granular, verifiable carbon accounting as part of environmental, social, and governance (ESG) assessments~\cite{esg1, esg2, esg3}.

These regulations are based on the GHG Protocol Corporate Standard, which defines emissions across three scopes: Scope~1 direct emissions, Scope~2 indirect emissions from electricity consumption, and Scope~3 all other indirect emissions across the value chain~\cite{Scope_3}. Among these, more than 75\% GHG emission reported are Scope~3 emissions~\cite{stinchcombe2023assessing, luo2012corporate}. However, they are particularly challenging to estimate, as they involve emissions from upstream and downstream activities such as purchased goods, transportation, and waste disposal. Making it necessary to investigate the enterprise value chain and accurately identify company sector categories~\cite{sectorclassify1, sectorclassify2, sectorclassify3}. Moreover, over 70\% of enterprises' emission reporting relies on Scope 3 emission factors, which are expensive to access and require domain expertise to determine carbon intensity~\cite{ExioML, dumit2024atlas}. This complexity hinders broader adoption, especially among small and medium enterprises (SMEs), and is exacerbated by a dearth of large-scale benchmark datasets that automate sector classification and carbon factor assignment~\cite{afolabi2023exploration}. The absence of such resources creates a major barrier, limiting GHG reporting largely to organizations with sufficient capital and expertise.

To address these challenges, we introduce \textbf{G}roup \textbf{R}easoning \textbf{E}mission \textbf{E}stimation \textbf{N}etworks (\textbf{GREEN}), the first LLM-driven enterprise emission estimation framework in an end-to-end manner. The predicted emission is the multiplication of an enterprise’s annual revenue and a carbon intensity factor, determined by classifying the enterprise into a sector. We fine-tune Sentence-BERT models via self-supervised contrastive learning and apply a \emph{Group Reasoning} hierarchical search with LLMs. Trained on a large-scale benchmark dataset constructed from scratch, named \emph{ExioNAICS}, it covers over 20,850 enterprises, each mapped to validated North American Industry Classification System (NAICS) codes~\cite{murphy1998introducing}. The Scope~3 emission factors are obtained from the ExioML economic dataset~\cite{ExioML} over 166 sectors. We formulate and standardize the automated emission estimation pipeline as an Information Retrieval (IR) problem and demonstrate the potential of Natural Language Processing (NLP) techniques in streamlining carbon accounting. We achieve 83.68\% Top-1 accuracy and 91.47\% Top-10 accuracy in the challenging \emph{industry classification} with 1,114 categories. The predicted enterprise emissions are evaluated with self-disclosed emissions found through sustainability reports, showing a moderate percentage error of 45.88\% on average.

This study contributes in three key ways:
\begin{enumerate}
\item It provides a standardized emission estimation pipeline that helps bridge machine learning research with climate science, thereby making carbon accounting more accessible to SMEs.
\item It introduces a novel, publicly available benchmark dataset for enterprise-level GHG estimation with a unifying NAICS and emission factor database, and applies state-of-the-art NLP models to automate sector classification.
\item It proposes high-performance fine-tuning via self-supervised contrastive learning and \emph{Group Reasoning} search.
\end{enumerate}

\begin{figure}[t!]
    \centering
    \includegraphics[width=1\linewidth]{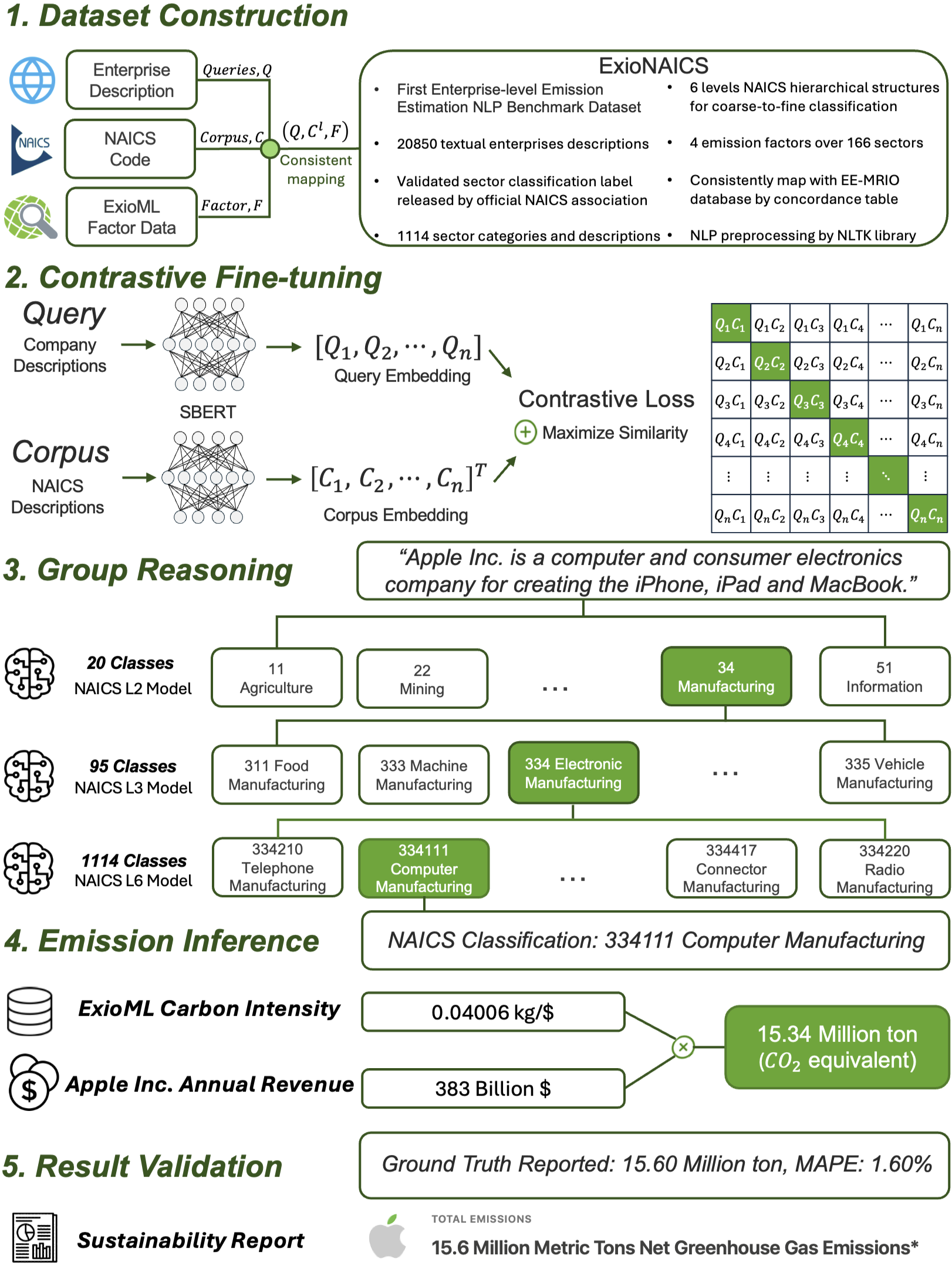}
    \caption{\textbf{High-level overview of the GREEN framework for enterprise-level emission estimation.} 
    (1)~We construct \emph{ExioNAICS}, a large-scale dataset pairing enterprise descriptions with validated NAICS codes and emission-factor data.
    (2)~A \emph{contrastive} fine-tuning step aligns embeddings of enterprise and NAICS descriptions.
    (3)~A \emph{Group Reasoning} approach hierarchically classifies each enterprise into a fine-grained NAICS sector. 
    (4)~Emission is inferred by multiplying the annual revenue by the sector’s carbon intensity. 
    (5)~Validation against official sustainability reports reveals small errors for single-sector firms and larger errors for diversified companies.}
    \label{fig:overview}
\end{figure}

\section{Related Work}
\subsection{Machine Learning in Sector Classification}
Machine Learning (ML) methods have been extensively explored for automating sector classification, a task traditionally reliant on expert-based taxonomies (e.g., GICS, NAICS). In a typical setup, each sample \(x_i\) (e.g., firm-level features, textual descriptions, or both) is mapped to a label \(y_i\) from a predefined category set \(\mathcal{Y}\). One seeks a classifier
\[
    f_{\theta}: X \to y,
\]
parameterized by \(\theta\). Minimizing a suitable loss, such as
\[
    \hat{\theta} \;=\; \arg\min_{\theta} \; \frac{1}{N}\sum_{i=1}^{N} \ell\bigl(f_{\theta}(x_i),\, y_i\bigr) \;+\; \lambda\,\Omega(\theta),
\]
lies at the core of traditional supervised learning. However, purely human-assigned labels face key obstacles: inconsistent coding across experts~\cite{sylolypavan2023impact}, limited coverage of new or cross-sector activities, poor scalability, and high annotation costs~\cite{sectorclassify9, sectorclassify10}. These limitations motivate automated, data-driven approaches.

Efforts to automate sector classification have evolved through three main stages. \textbf{Stage~I: Traditional ML on Tabular Data.} Early work leveraged structured firm attributes (e.g., financial statements) with models like Random Forests, K-Nearest Neighbors, and SVMs~\cite{sectorclassify4, sectorclassify5, sectorclassify6, sectorclassify9, sectorclassify12, sectorclassify1}, but often faced small datasets, domain shifts, and limited representational power. \textbf{Stage~II: Text-Based Frequency Models.} With growing availability of unstructured data (e.g., 10-K reports, descriptions), researchers used Bag-of-Words or TF-IDF transformations fed into classifiers like MLPs~\cite{sectorclassify4, sectorclassify5, sectorclassify8, sectorclassify7, sectorclassify11}. However, they still struggled with shallow context and large label spaces. \textbf{Stage~III: Transformer-Based LLMs.} Modern approaches employ pre-trained Transformers such as BERT or Sentence-BERT (SBERT), which encode deeper semantics and demonstrate strong zero-shot or fine-tuned performance~\cite{jain2024empowering, balaji2023flamingo, balaji2023caml, sectorclassify3}. These methods surpass older models but require large-scale computing, careful domain adaptation, and open-source data to maintain reproducibility.

While LLMs offer state-of-the-art accuracy, challenges persist around dataset openness, computational demands, and extending classification beyond narrow taxonomies toward broader tasks like emission estimation. Future advances in flexible, interpretable, and efficient LLMs will be critical for real-world industrial applications.

\subsection{Self-Supervised Contrastive Learning Framework}
Self-supervised learning (SSL) has emerged as a powerful representation learning paradigm that does not require large labeled datasets. Instead, the model learns from inherent data structures, creating \emph{positive} and \emph{negative} instances by various transformations or pairing strategies. SSL shifts away from cross-entropy on labeled samples \(\{(x_i,y_i)\}\) and instead uses contrastive losses to align similar views of the same data point while separating different samples.

Initial advances in SSL stemmed from the image domain, with frameworks like SimCLR~\cite{SimCLR} and MoCo~\cite{MoCo} leveraging an InfoNCE loss to bring positive pairs (augmented views of the same image) closer in latent space relative to a set of negatives. Later works like BYOL~\cite{contractiveBYOL} and SimSiam~\cite{Siamese} showed negative-free designs. In NLP, models such as SBERT~\cite{reimers2019sentence} and SimCSE~\cite{gao2021simcse} adapted contrastive principles to sentence embeddings, enabling robust similarity measures with minimal or no labeled data. Contrastive methods have thus evolved into a general framework for embedding diverse data types (images, text, multimodal) into semantically meaningful spaces.

\subsection{GHG Emission Estimation by Ecological Economic Framework}
Over 70\% of enterprises estimate their carbon footprints using \emph{sector-based carbon intensity factors}, representing GHG emissions produced per unit of economic output in a given sector and region~\cite{dumit2024atlas}. The Environmentally Extended Multi-Regional Input–Output (EE-MRIO) framework provides a structured way to derive these intensities by integrating economic transactions and regional environmental data~\cite{wiedmann2009review, leontief1963multiregional}. The carbon intensity factor is defined as the ratio of a sector’s total emissions to its economic output. While the EE-MRIO framework offers a comprehensive view of inter-sector linkages, its deployment in real industrial applications is hindered by expensive data and domain expertise requirements~\cite{dietzenbacher2013construction, aguiar2016overview}.

\section{Method}
\subsection{Open-Source Large-Scale NLP Benchmark Dataset: \emph{ExioNAICS}}
Despite recent advancements in LLMs, sector classification and emission estimation still lack publicly available, large-scale datasets. Existing work often uses closed-source repositories or small label spaces. We therefore introduce \emph{ExioNAICS}, the first open-source dataset targeting \emph{both} sector classification and emission estimation. It integrates:
\begin{itemize}
\item \textbf{NAICS Codes and Descriptions.} We adopt the North American Industry Classification System (NAICS). Validated NAICS codes are retrieved from the official NAICS Association, mitigating label noise.
\item \textbf{EE-MRIO Emission Factors.} We link NAICS sectors to the \emph{ExioML} database~\cite{ExioML}, an open-source extension of EXIOBASE~\cite{Exiobase}, containing multi-regional input–output tables and environmental data across 49 regions and 163 sectors (1995–2022).
\end{itemize}

We gathered over 20,850 textual descriptions from 13,823 unique enterprises covering diverse sectors. The dataset preserves NAICS’s hierarchical granularity: codes at the 2-digit level define 20 broad categories, while 6-digit codes define 1,114 specialized categories. Table~\ref{tab:dataset_statistics} shows key statistics. We unify these textual data with carbon intensity factors from ExioML, effectively bridging the gap between text-based classification and numerical emission-factor assignment.

\begin{table}[h!]
    \centering
    \caption{Key Statistics of the ExioNAICS Dataset}
    \label{tab:dataset_statistics}
    \begin{tabular}{l||c|c|c|c|c}
    \toprule
    \textbf{Metric} & \(\;Q\) & \(\;C^2\) & \(\;C^3\) & \(\;C^6\) & \(\;F\) \\
    \midrule
    Unique Class & 13,823 & 20 & 95 & 1,114 & 119 \\
    Min Length   & 6      & 103 & 17 & 10   & --  \\
    Avg Length   & 33     & 515 & 153 & 56   & --  \\
    Max Length   & 154    & 846 & 430 & 164  & --  \\
    \midrule
    \multicolumn{5}{l}{\textbf{Data Size} \( \vert D \vert\)} & 20,850 \\
    \bottomrule
    \end{tabular}
\end{table}

\subsection{Sentence-BERT with Contrastive Fine-Tuning}
We cast sector classification as an Information Retrieval (IR) problem: for a given enterprise description \( q \), retrieve the most relevant NAICS document \( c \in C^l \). This approach naturally scales to large or evolving taxonomies, unlike standard classification with a rigid label space.

We adopt the Sentence-BERT (SBERT) framework~\cite{reimers2019sentence}, which uses a \emph{siamese} encoder to produce fixed-dimensional sentence embeddings. Let \(f_{\theta}: X \mapsto \mathbb{R}^d\) be the shared encoder. For each query \(q\) and NAICS document \(c\), the embeddings are:
\[
  Q = f_{\theta}(q), 
  \quad
  C = f_{\theta}(c).
\]
Their similarity is measured by cosine similarity
\[
s(q,c) = \frac{Q \cdot C}{\|Q\|\;\|C\|}.
\]
Hence, the most relevant NAICS class is found by Maximum Inner Product Search (MIPS):
\[
    \pi(q) = \arg\max_{c \in C^l} \; s(q,c).
\]

Rather than a standard cross-entropy, we fine-tune SBERT using a \emph{contrastive} Multiple Negative Ranking (MNR) loss~\cite{MNRloss}:
\[
    \mathcal{L}
    \;=\;
    -\frac{1}{n}\,\sum_{i=1}^{n}\,
    \log\Bigl(
    \frac{\exp(\cos(Q_i, D_i))}
         {\sum_{j=1}^{n}\,\exp(\cos(Q_i, D_j))}
    \Bigr),
\]
where \(Q_i=f_{\theta}(q_i)\) and \(D_i=f_{\theta}(d_i)\) are positive query--document pairs, while all other documents in the same batch serve as negatives. This encourages alignment of relevant pairs and separation from non-relevant pairs.

\begin{figure}[t!]
    \centering
    \includegraphics[width=0.4\linewidth]{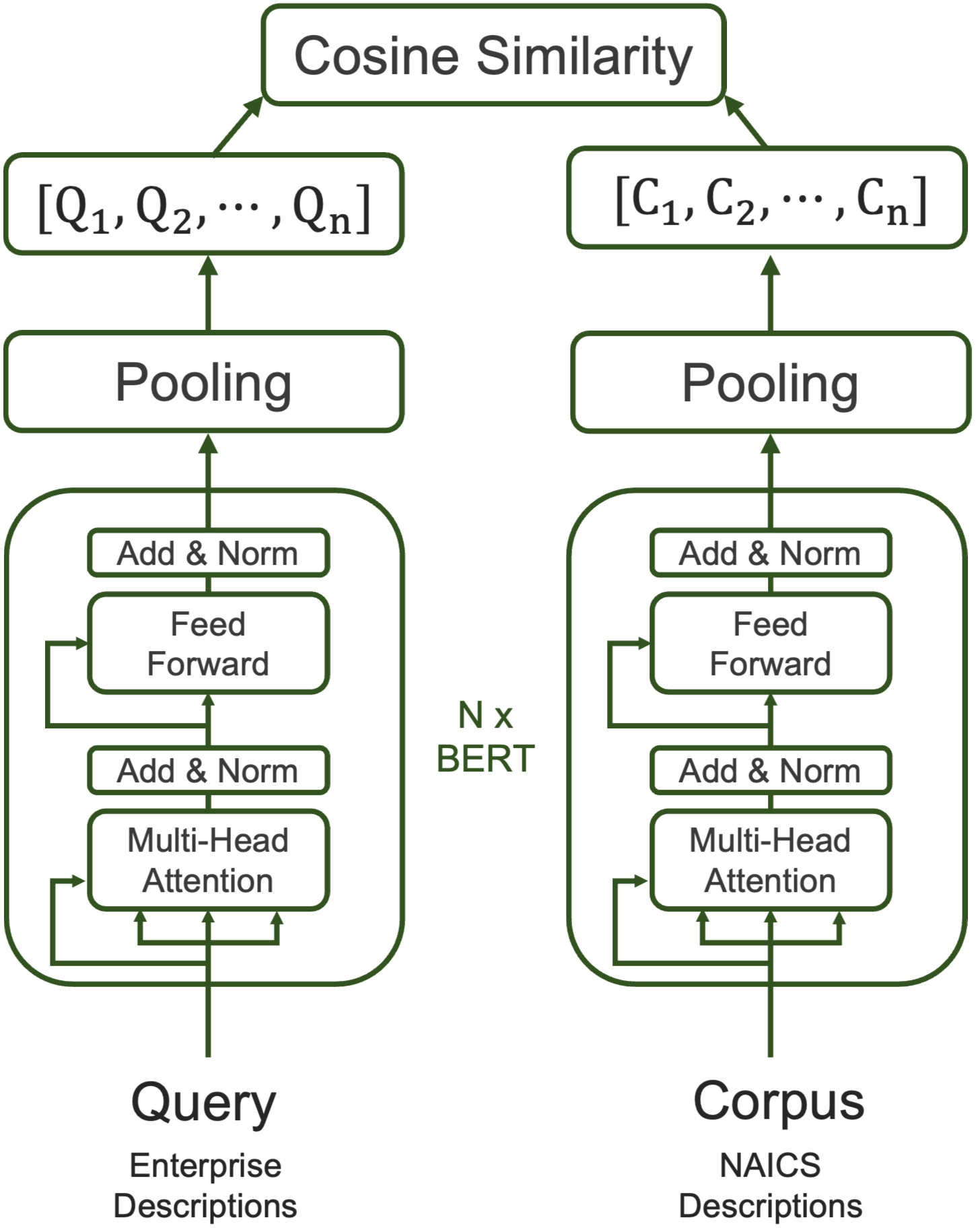}
    \caption{\textbf{Architecture of Sentence-BERT (SBERT)} for enterprise classification as an IR task. 
    The model encodes both \emph{queries} (enterprise descriptions) and \emph{corpus} documents (NAICS definitions) into a shared embedding space, 
    where cosine similarity measures relevance. 
    Fine-tuning uses a contrastive loss separating correct from incorrect matches.}
    \label{fig:sbert}
\end{figure}

\subsection{Large-Scale Sector Classification via Hierarchical \emph{Group Reasoning}}
NAICS codes have a \emph{coarse-to-fine hierarchy}: 20 categories at level~2, 95 at level~3, and 1,114 at level~6. Classifying queries directly among 1,114 labels is complex, and single-stage classifiers often suffer from higher uncertainty and heavier computation. We introduce a \emph{Hierarchical Group Reasoning} method, which ensembles multiple LLM-based classifiers and domain-specific heuristics at each level. This approach:
\begin{enumerate}
    \item Decomposes the large classification task into smaller, level-wise subproblems.
    \item Traverses the NAICS tree from root to leaves, pruning irrelevant branches early.
    \item Reduces model uncertainty (entropy) and lowers time complexity from exponential (\(b^d\)) to linear in the depth (\(b \cdot d\)).
\end{enumerate}

Algorithm~\ref{alg:group_reasoning} details the procedure. A top-$k$ parameter controls how many child nodes at each level are selected for expansion, trading off accuracy and speed. Theoretical proofs show that hierarchical decomposition lowers Shannon entropy compared to a single massive classifier and cuts computational overhead~(\S\ref{sec:theory}).

\begin{algorithm}[t!]
\caption{Hierarchical Group Reasoning of LLMs}
\label{alg:group_reasoning}
\begin{algorithmic}[1]
\Require 
  A hierarchical taxonomy $\mathcal{H}$ with $L$ levels; 
  level-specific LLMs $\{f_{\theta_\ell}\}_{\ell=1}^{L}$; 
  a test query $q$; 
  a root node $r$;
  top-$k$ selection parameter $k$.
\Ensure 
  Predicted fine-grained NAICS class $\hat{y}$ at level $L$.

\vspace{1mm}
\State $N \gets \{r\}$ \Comment{Start from the root node}
\For{$\ell \gets 1$ to $L$}
    \State $Q \gets f_{\theta_\ell}(q)$ \Comment{Embed the query at level $\ell$}
    \State $\mathcal{S} \gets \emptyset$ 
    \For{each node $n \in N$}
        \State $\mathcal{C}_n \gets \text{Children}(n)$ 
        \For{each child node $c \in \mathcal{C}_n$}
            \State $C \gets f_{\theta_\ell}(c)$ \Comment{Embed the child document}
            \State $s \gets \cos\bigl(Q,\,C\bigr)$ 
            \State $\mathcal{S} \gets \mathcal{S} \cup \{(c, s)\}$ 
        \EndFor
    \EndFor
    \State $N \gets \text{Top-}k(\mathcal{S}, k)$ \Comment{Pick $k$ nodes to expand}
\EndFor
\State \textbf{return} $\hat{y} \gets \arg\max_{(c,s)\in\mathcal{S}} s$ 
\end{algorithmic}
\end{algorithm}

\subsection{Theoretical Performance Analysis}
\label{sec:theory}
\begin{theorem}[Hierarchical Classification Entropy]\label{theorem_1}
Let there be a hierarchical classification tree of depth \(d\) with uniform branching \(b\). Let each level \(i\) have accuracy \(p_i\). The hierarchical approach has strictly lower Shannon entropy than a single-stage classifier with \(b^d\) classes and accuracy \(\prod_{i=1}^d p_i\). Formally, \(H_{D}(Y) \ge H_G(Y)\).
\end{theorem}

\begin{theorem}[Complexity]\label{theorem_2}
A single-stage approach over \(b^d\) classes has \(O(b^d)\) time, whereas hierarchical classification has \(O(b\cdot d)\).
\end{theorem}

\section{Experiments \& Results}
\subsection{Experimental Setup}
We evaluate on NAICS levels~2 (20 classes), 3 (95 classes), and 6 (1,114 classes). Data splits are 80\% train, 10\% validation, 10\% test. Models are trained for 100 epochs with a learning rate \(2\times10^{-5}\) on an NVIDIA T4 GPU. The primary metric is Top-$k$ accuracy (\(\text{Acc}@k\)):
\[
    \text{Acc}@k 
    = 
    \frac{\text{TP}}{\text{TP} + \text{FP} + \text{FN}},
\]
where a prediction is correct if \emph{any} ground-truth label is among the top-$k$ predictions. We examine the effects of different SBERT backbones, data augmentation, hierarchical hyperparameters, and ensemble strategies.

\subsection{Comparison of Pre-trained SBERT Backbones}
Table~\ref{tab:model_performance} compares multiple Sentence-Transformer (SBERT) backbones on levels 2, 3, and 6. These differ in model size (60\,MB to 420\,MB) and pre-training corpora (\emph{Paraphrase}, \emph{Multi-QA}, \emph{All}).

\begin{table}[h!]
    \centering
    \caption{\textbf{Top-1 Accuracy} of various SBERT Backbones on NAICS-2, -3, -6 tasks}
    \label{tab:model_performance}
    \begin{tabular}{l||c|c|c|c|c}
    \toprule
    \textbf{Backbone} & \textbf{Pre-train} & \textbf{Size (MB)} & \(\;C^2\)&\(\;C^3\) & \(\; C^6 \)\\
    \midrule
    MiniLM$_{L3}$ & Paraphrase & 60  & 89.73  & 86.02  & 77.51 \\
    MiniLM$_{L6}$ & Multi-QA   & 80  & 89.14  & 86.52  & 79.57\\
    MiniLM$_{L6}$ & All        & 80  & 91.68  & 87.93  & 78.34 \\
    MiniLM$_{L12}$& All        &120  & 91.23  & 87.84  & 82.67\\
    Mpnet$_{base}$& All        &420  & \textbf{91.73} & \textbf{88.54} & \textbf{82.87} \\
    \hline
    \end{tabular}
\end{table}

\emph{Mpnet$_{base}$} achieves the highest accuracy at all levels, but it is quite large (420\,MB) and slow. Smaller \emph{MiniLM} variants remain competitive, e.g., \emph{MiniLM$_{L12}$} is only 120\,MB with near state-of-the-art performance. Hence, practitioners can balance performance against computational constraints.

\subsection{Data Augmentation Analysis}
We test paraphrase-based augmentations (e.g., word replacement) at random probabilities for \emph{MiniLM$_{L3}$} on NAICS-6. Table~\ref{tab:replacement_performance} shows minimal gains or slight performance drops, consistent with findings that large pre-trained models are robust to naive text augmentations.

\begin{table}[h!]
    \centering
    \caption{\textbf{Accuracy} (\%) with different data augmentations (NAICS-6, 1,114 classes).}
    \label{tab:replacement_performance}
    \begin{tabular}{l||c|c|c|c}
    \toprule
    \textbf{Method} & \(\text{Acc}@1\) & \(\text{Acc}@3\) & \(\text{Acc}@5\) & \(\text{Acc}@10\) \\
    \midrule
    Non-augmented & 77.51 & \textbf{87.37} & \textbf{89.22} & \textbf{91.33} \\
    BERT Replace  & 77.06 & 86.18 & 87.77 & 89.87 \\
    Noun Replace  & \textbf{77.93} & 85.31 & 87.77 & 90.74 \\
    Verb Replace  & 76.77 & 85.53 & 87.19 & 89.94 \\
    Adj/Adv Replace & 76.63 & 85.46 & 87.48 & 90.16 \\
    Random Replace& 75.83 & 85.53 & 87.41 & 90.38 \\
    \bottomrule
    \end{tabular}
\end{table}

\subsection{Group Reasoning Hyperparameter $k$}
Table~\ref{tab:accuracy_time_comparison} shows that increasing $k$ (the beam width in hierarchical search) yields higher accuracy but linearly increases running time. For $k=90$, \(\text{Acc}@1\) rises from 77.51\% to 80.29\% with a total runtime of 16 minutes, still faster than naive MIPS over 1,114 classes.

\begin{table}[h!]
    \centering
    \caption{\textbf{Accuracy} and inference time for varying $k$ (NAICS-6, 1,114 classes).}
    \label{tab:accuracy_time_comparison}
    \begin{tabular}{l||c|c|c|c|c}
    \toprule
    Method & \(\text{Acc}@1\) & \(\text{Acc}@3\) & \(\text{Acc}@5\) & \(\text{Acc}@10\) & Time (min) \\
    \midrule
    MIPS (baseline) & 77.51 & 87.37 & 89.22 & 91.33 & 15 \\
    $k=10$ & 76.32 & 84.63 & 86.84 & 88.38 & \textbf{7} \\
    $k=20$ & 78.46 & 87.13 & 89.12 & 90.29 & 7 \\
    $k=30$ & 78.97 & 87.43 & 89.93 & 90.81 & 8 \\
    $k=40$ & 79.41 & 87.57 & 89.93 & 90.81 & 10 \\
    $k=50$ & 79.56 & 87.79 & 90.15 & 90.96 & 11 \\
    $k=60$ & 79.78 & 87.94 & 90.22 & 91.10 & 12 \\
    $k=70$ & 79.93 & 88.16 & 90.22 & 91.25 & 14 \\
    $k=80$ & 80.07 & 88.31 & \textbf{90.37} & 91.47 & 16 \\
    $k=90$ & \textbf{80.29} & \textbf{88.31} & 90.29 & \textbf{91.54} & 16 \\
    \bottomrule
    \end{tabular}
\end{table}

\subsection{Model Ensembles in Group Reasoning}
We can ensemble smaller and larger \emph{MiniLM} backbones at different levels. For instance, use L3 at levels~2 and 3, then L12 at level~6. Table~\ref{tab:accuracy_comparison} shows that mixing smaller models for early levels and bigger ones for deeper levels can achieve high accuracy with moderate size.

\begin{table}[h!]
    \centering
    \caption{\textbf{Accuracy} (\(\text{Acc}@k\)) for various ensemble configurations in Group Reasoning (NAICS-6).}
    \label{tab:accuracy_comparison}
    \begin{tabular}{l||c|c|c|c|c}
    \toprule
    \textbf{Model Ensemble} & \textbf{Size (MB)} & \(\text{Acc}@1\)&\(\text{Acc}@3\)&\(\text{Acc}@5\)&\(\text{Acc}@10\)\\
    \midrule
    MiniLM$_{L3}$ (single) & \textbf{60}  & 77.51 & 87.37 & 89.22 & 91.33\\
    MiniLM$_{L3,3,3}$       & 180         & 80.29 & 88.31 & 90.29 & 91.54 \\
    MiniLM$_{L3,3,6}$       & 200         & 80.44 & \textbf{89.04} & \textbf{91.18} & \textbf{92.94}\\
    MiniLM$_{L3,3,12}$      & 240         & 83.24 & 88.60 & 90.44 & 91.54\\
    MiniLM$_{L3,6,12}$      & 260         & \textbf{83.68} & 88.60 & 90.44 & 91.47\\
    \bottomrule
    \end{tabular}
\end{table}

\subsection{Ablation Study}
Table~\ref{tab:retrieval_performance} shows incremental ablations for NAICS-6 with \emph{MiniLM$_{L3}$}. Zero-shot SBERT yields only 20.12\% \(\text{Acc}@1\). Cross-entropy slightly improves to 21.49\%, but contrastive MNR drastically jumps to 76.85\%. NLTK preprocessing, Group Reasoning, and the multi-level ensemble lead to the final \textbf{GREEN} model with 83.68\%.

\begin{table}[h!]
    \centering
    \caption{\textbf{Ablation Study} (NAICS-6, 1,114 classes) using \emph{MiniLM$_{L3}$} unless noted.}
    \label{tab:retrieval_performance}
    \begin{tabular}{l||c|c|c|c}
    \toprule
    \textbf{Method} & \(\text{Acc}@1\)&\(\text{Acc}@3\)&\(\text{Acc}@5\)&\(\text{Acc}@10\) \\
    \midrule
    SBERT Zero-shot     & 20.12 & 35.51 & 43.27 & 53.52 \\
    SBERT + CE          & 21.49 & 36.39 & 45.58 & 55.42 \\
    SBERT + MNR         & 76.85 & 85.42 & 87.85 & 90.28 \\
    + NLTK Preprocess   & 77.51 & 87.37 & 89.22 & 91.33 \\
    + Data Augmentation & 77.93 & 85.31 & 87.77 & 90.73 \\
    + Group Reasoning   & 80.01 & 88.31 & 90.37 & 91.47 \\
    \midrule
    \textbf{GREEN} (\emph{Multi-level}) & \textbf{83.68} & \textbf{88.60} & \textbf{90.44} & \textbf{91.47}\\
    \bottomrule
    \end{tabular}
\end{table}

\section{Enterprise Emission Inference}
We estimate corporate GHG emissions by multiplying predicted carbon intensity from NAICS classification with annual revenue. Due to limited public data, we sample 20 companies with \emph{self-disclosed} emissions, compare with GREEN estimates, and compute Mean Absolute Percentage Error (MAPE):
\[
\text{MAPE} = \frac{100\%}{N} \sum_{i=1}^{N} \left| \frac{R_i - E_i}{R_i} \right|,
\]
where \(R_i\) is reported emissions, \(E_i\) is the GREEN estimate.

Table~\ref{tab:ghg_predictions} shows results. The overall MAPE is 45.88\%. Single-sector companies like Apple and Air Canada exhibit lower errors, while diversified giants (e.g., Amazon, Samsung) show larger discrepancies. Errors stem from (1) factor bias in EE-MRIO, (2) single-sector misclassifications, or (3) cross-sector complexities not captured by one label. Future improvements include multi-sector classification and refined emission-factor modeling. Nonetheless, \emph{GREEN} provides a practical and transparent baseline for enterprise-level GHG reporting.

\begin{table}[h!]
    \centering
    \caption{Comparison of GREEN-Estimated vs.\ Self-Disclosed GHG Emissions for 20 Companies.}
    \label{tab:ghg_predictions}
    \small
    \begin{tabular}{l||c|c|c|c|c|c}
    \toprule
    \textbf{Company} & \textbf{Revenue} & \textbf{Intensity} & \textbf{Estimated} & \textbf{Reported} & \textbf{MAPE} & \textbf{Err. Type} \\
    \midrule
    Apple   & 383.00 & 0.0388 & 15.34   & 15.60   & 1.60   & Factor bias \\
    John Deere   & 15.50  & 5.7155 & 91.25   & 97.00   & 5.93   & Misclassified \\
    Air Canada    & 12.74  & 1.6340 & 21.44   & 19.63   & 9.23   & Factor bias \\
    Tencent       & 86.00  & 0.0588 & 5.06    & 5.79    & 10.04  & Factor bias \\
    Google       & 305.63 & 0.0388 & 11.89   & 14.30   & 16.87  & Factor bias \\
    Microsoft   & 227.58 & 0.0865 & 20.28   & 17.16   & 18.22  & Misclassified \\
    Telsa      & 96.77  & 0.3988 & 39.75   & 48.91   & 18.72  & Factor bias \\
    Nike       & 49.10  & 0.1446 & 7.32    & 10.03   & 27.05  & Factor bias \\
    NVIDIA      & 26.97  & 0.0572 & 1.59    & 2.24    & 29.06  & Factor bias \\
    Meta         & 131.90 & 0.0865 & 11.78   & 8.45    & 39.14  & Factor bias \\
    Murphy Oil    & 3.46   & 3.3522 & 11.95   & 24.30   & 50.84  & Cross sector \\
    ADM           & 93.00  & 1.7791 &170.42   &107.00   & 59.27  & Cross sector \\
    Dole           & 8.00   & 0.3052 & 2.52    & 7.00    & 64.07  & Cross sector \\
    Shell        &381.31  & 1.8691 &734.10   &2048.00  & 64.16  & Cross sector \\
    Toyota         &307.00  & 0.3988 &126.11   &575.73   & 78.09  & Cross sector \\
    Vinci         & 49.40  & 0.0538 & 2.74    & 12.98   & 78.89  & Cross sector \\
    Zentis       & 0.46   & 0.2207 & 0.11    & 0.72    & 85.29  & Cross sector \\
    Amazon     &574.00  & 0.0388 & 23.03   & 68.82   & 90.65  & Cross sector \\
    Samsung     &194.00  & 0.0444 & 8.88    &124.72   & 92.88  & Cross sector \\
    FedEx     & 90.40  & 0.0008 & 0.08    & 22.98   & 99.66  & Cross sector \\
    \midrule
    \textbf{Average MAPE (\%)} & \multicolumn{6}{c}{\textbf{45.88}} \\
    \bottomrule
    \end{tabular}
\end{table}

\section{Conclusion}
We propose \textbf{GREEN} as the first end-to-end LLM-driven framework for \emph{enterprise-level} GHG emission estimation. The core pipeline: 
(1)~maps textual descriptions to NAICS sectors via \emph{contrastive} SBERT classification, 
(2)~derives carbon intensity factors from ExioML, and 
(3)~computes emissions as revenue \(\times\) intensity. 
We introduce \emph{ExioNAICS}, a large-scale public benchmark unifying 20,850 enterprises across 1,114 NAICS categories with emission-factor data, and propose a novel \emph{Group Reasoning} method to handle large-scale hierarchical classification efficiently. 
Extensive experiments show \(\text{Acc}@1=83.68\%\) on NAICS-6, surpassing prior methods, and a moderate MAPE of 45.88\% when validated against self-disclosed corporate emissions.

\clearpage
\bibliography{bib}
\bibliographystyle{ieeetr}
\end{document}